\documentclass[lettersize,journal]{IEEEtran}
\usepackage{amsmath,amsfonts}
\usepackage{algorithmic}
\usepackage{array}
\usepackage[caption=false,font=normalsize,labelfont=sf,textfont=sf]{subfig}
\usepackage{textcomp}
\usepackage{stfloats}
\usepackage{url}
\usepackage{verbatim}
\usepackage{graphicx}
\usepackage{booktabs}
\usepackage{makecell} 
\usepackage{float}
\usepackage{amssymb}
\usepackage{hyperref}
\usepackage{mathtools}
\usepackage{bm}
\def\BibTeX{{\rm B\kern-.05em{\sc i\kern-.025em b}\kern-.08em
    T\kern-.1667em\lower.7ex\hbox{E}\kern-.125emX}}
\usepackage{balance}
\newcommand{\Rom}[1]{\uppercase\expandafter{\romannumeral#1\relax}}
\begin{document}
\title{A Tactile-based Interactive Motion Planner for Robots in Unknown Cluttered Environments}
\author{
    Chengjin Wang,~\IEEEmembership{Student Member,~IEEE,}
    Yanmin Zhou,~\IEEEmembership{Member,~IEEE,}
    Zheng Yan, 
    Feng Luan,  
    Runjie Shen, 
    Hongrui Sang, 
    Zhipeng Wang, 
    Bin He,~\IEEEmembership{Senior Member,~IEEE}

\thanks{This work was supported in part by the National Natural Science Foundation of China (No. 62088101), the Science and Technology Commission of Shanghai Municipality (No.2021SHZDZX0100, 22ZR1467100), and the Fundamental Research Funds for the Central Universities (No.22120240291). (Corresponding author: Yanmin Zhou). 

Chengjin Wang, Zheng Yan and Feng Luan are with the Shanghai Research Institute for Intelligent Autonomous Systems Shanghai 201210, China, with the State Key Laboratory of Autonomous Intelligent Unmanned Systems Shanghai 201210, China, and with the Frontiers Science Center for Intelligent Autonomous Systems, Shanghai 201210, China. 
Yanmin Zhou, Runjie Shen, Hongrui Sang, Zhipeng Wang, and Bin He are with the College of Electronics and Information Engineering, Tongji University, Shanghai 201804, China, with the state Key Laboratory of Autonomous Intelligent Unmanned Systems, Shanghai 201210, China, and with Frontiers Science Center for Intelligent Autonomous Systems, Shanghai 201210, China.

Additional videos in this article are available online at  \protect\url{https://travelers-lab.github.io/I-MP/}}}

\markboth{IEEE TRANSACTIONS ON CYBERNETIC}%
{A Tactile-based Interactive Motion Planner for Robots in Unknown Cluttered Environments}

\maketitle

\begin{abstract}
In unknown cluttered environments with densely stacked objects, the free-motion space is extremely barren, posing significant challenges to motion planners. Collision-free planning methods often suffer from catastrophic failures due to unexpected collisions and motion obstructions. To address this issue, this paper proposes an interactive motion planning framework (I-MP), based on a perception-motion loop. This framework empowers robots to autonomously model and reason about contact models, which in turn enables safe expansion of the free-motion space. Specifically, the robot utilizes multimodal tactile perception to acquire stimulus-response signal pairs. This enables real-time identification of objects' mechanical properties and the subsequent construction of contact models. These models are integrated as computational constraints into a reactive planner. Based on fixed-point theorems, the planner computes the spatial state toward the target in real time, thus avoiding the computational burden associated with extrapolating on high-dimensional interaction models. Furthermore, high-dimensional interaction features are linearly superposed in Cartesian space in the form of energy, and the controller achieves trajectory tracking by solving the energy gradient from the current state to the planned state. The experimental results showed that at cruising speeds ranging from 0.01 to 0.07 $m/s$, the robot's initial contact force with objects remained stable at 1.0$\pm$0.7 $N$. In the cabinet scenario test where collision-free trajectories were unavailable, I-MP expanded the free motion space by 37.5\% through active interaction, successfully completing the environmental exploration task.
\end{abstract}

\begin{IEEEkeywords}
Unknown cluttered environments, interactive motion planning, electronic skin, contact modeling, 
\end{IEEEkeywords}

\section{Introduction}
\label{sec_1}
\IEEEPARstart{T}{he} capability of a planner to overcome the unsolvable free-motion space enables robots to explore potential motion paths in cluttered environments where collision-free trajectories are unavailable \cite{doi:10.1126/scirobotics.adf7843, doi:10.1177/0278364912471865, scordamaglia2025autonomous, 8954627, aucone2024synergistic}. Interactive motion planning, which actively reconfigure the spatial configuration of the environment to expand the free-motion space, has emerged as a promising approach. However, in complex physical interactions influenced by high-dimensional environmental features, enabling robots to autonomously build and extrapolate contact models in real time remains a significant challenge \cite{doi:10.1177/0278364912471865}. Ideally, robots can adopt a closed-loop perception-action strategy: on one hand, autonomously constructing contact models through action-guided perception, and on the other hand, generating predictable interactive actions via perception-guided behavior.

\subsection{Literature Review}
\label{sec_1_1}
Given the advantages of interactive planners in addressing spatial configuration management, this paper summarizes the state-of-the-art advances in interactive motion planning and categorizes them according to three criteria: (a) computational constraints in planning, (b) predictability of planned movements, and (c) autonomous modeling of physical interactions, as shown in TABLE \ref{tab1}.
\begin{table}[t] 
    \centering
    \caption{Comparative Analysis of Planning Methods in Cluttered Environments} 
    \label{tab1} 
    
    \begin{tabular}{lccc}
        \toprule
        Algorithm & \makecell{Interactive\\Constraint} & Predictability & \makecell{Contact\\Modeling}\\
        \midrule
        \makecell[l]{\textbf{Artificial Potential} \\\textbf{Field} \cite{khatib1986real}} & - & - & -\\
        \textbf{Sampling-based} \cite{barraquand1991robot} & $\circ$ & - & -\\
        \makecell[l]{Adaptive Motion \\Primitive}\cite{brouwer2024tactile} & $\circ$ & - & - \\
        \makecell[l]{Vision-Language-\\Action} \cite{zitkovich2023rt} & $\circ$ & $\circ$ & - \\
        \makecell[l]{Finite Element \\Method} \cite{gansterer2014simulation} & $\bullet$ & $\bullet$ & - \\
        \makecell[l]{\textbf{Open Dynamic} \\\textbf{Engine}} \cite{10161006} & $\bullet$ & $\bullet$ & - \\
        Kinodynamic \cite{donald1993kinodynamic} & $\bullet$ & $\bullet$ & - \\
        \textbf{I-MP} & $\bullet$ & $\bullet$ & $\bullet$ \\
        \bottomrule
    \end{tabular}
    \vspace{0.2cm}
    \begin{minipage}{0.95\linewidth}

    \footnotesize The frameworks discussed in this study are written in bold text.\\
    -: no ability \\
    $\circ$ : potential ability without explicit results \\
    $\bullet$ : full ability \\
    \end{minipage}
\end{table}

These planning methods can be categorized into three types according to their computational principles: probability-based, simulation-based, and model-based approaches. Probability-based planning methods aims to find motion states with maximum connectivity probabilities \cite{xiang2024grasping,liu2022real, braglia2023online, tang2024obstacle}, whose combined use with compliant control strategies can avoid damaged collisions \cite{aucone2024synergistic, zitkovich2023rt, odhner2014compliant,zhou2024active,brooks2024video, hang2021manipulation}. Simulation-based and model-based motion planning methods are put forward as well \cite{gansterer2014simulation, sundaresan2022diffcloud, ren2023kinodynamic, dogar2012physics}to integrate interaction features into planning.

Simulation-based motion planners can predict interaction effects such as contact forces, deformations, and displacements, and predicted results help significantly reduce machining errors \cite{gansterer2014simulation} or facilitate the assembly of deformable bodies \cite{yoshida2015simulation}. Model-based planning methods, including model-predictive planning \cite{pang2023planning} and kinodynamic planning \cite{donald1993kinodynamic, ren2023kinodynamic}, formulate the planning problem as a convex optimization, thereby generating interactive motion with dynamic constraints. Challenges such as environmental modeling, computational efficiency, and conversion of non-convex-to-convex optimization arise as key research focuses. However, deploying simulation-based or model-based planning methods in unknown, cluttered environments confront environmental modeling issues as robots find it difficult to autonomously build precise interaction models like system-identification devices. Ideally, planners should autonomously extract key interaction features and introduce them into the planning process to generate real-time interactive motions. The features that significantly influence the object interaction process primarily include geometric features and implicit physical features. The latter encompass the material and dynamic properties of objects, which are generally difficult to perceive visually.

\subsection{Proposed Solution}
\label{sec_1_2}

To achieve this goal, this paper proposes a closed-loop \textbf{i}nteractive \textbf{m}otion \textbf{p}lanning framework (I-MP). By integrating kinesthetic perception \cite{sintov2023simple} with interactive reasoning, the framework enables safe robot interaction in unknown cluttered environments. Specifically, it leverages multimodal tactile perceptions from electronic skin and proprioceptive information to extract environmental interaction features and build contact models. Tactile signals distributed across the robot's body undergo spatiotemporal consistency calibration to form stimulus-response observation pairs. Subsequently, system identification methods are employed to estimate the environmental interaction features from these observation pairs \cite{wang2024trafficability}.

To overcome the computational burden introduced by high-dimensional contact models, the planner adopts a reactive planning strategy based on state-space rationality discrimination. This strategy utilizes fixed-point theorems to identify in real time the shortest safe spatial state toward the target, thereby avoiding the computational cost of extrapolating models. To ensure safe tracking of the planned states, the contact model is linearly superimposed in Cartesian space in the form of energy, and the controller completes the control task by computing the energy gradient from the current state to the planned state. Furthermore, to reconcile the inherent conflict between kinesthetic perception and interactive motion, perception confidence is introduced to determine motion intentions, thereby constructing a fully closed-loop motion planning framework in which perception guides action and action guides perception. 

\begin{figure*}[!t]
\centering
\includegraphics[width= 0.8 \textwidth]{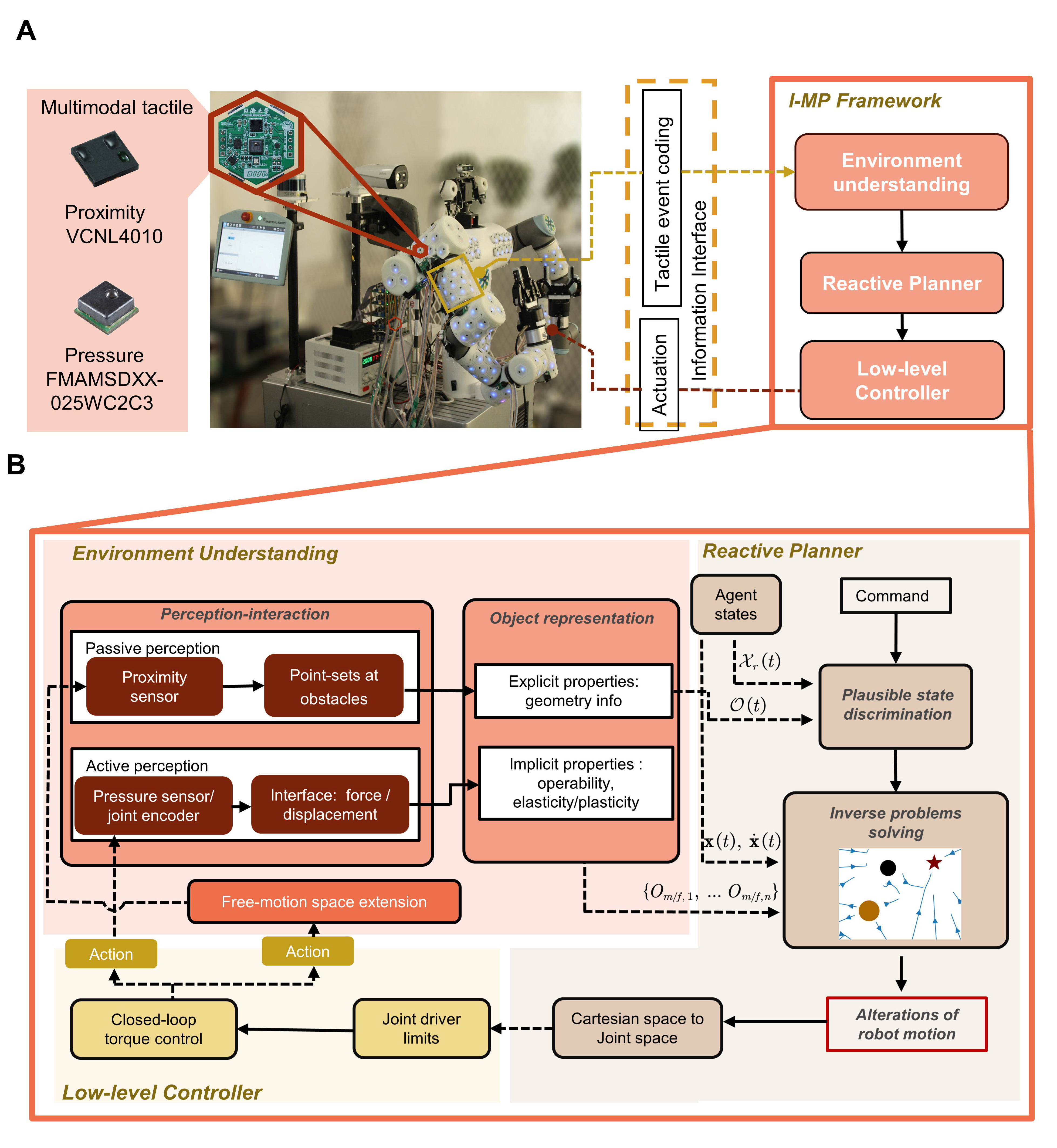} 
\caption{Overview of the I-MP Framework. (A) Overview of the motion planning architecture. (B) Core I-MP components. The I-MP consists of environment understanding (EU), reactive planner (RP), and low-level controller (LC) as its modules.}
\label{fig:1}
\end{figure*}

The specific implementation of the proposed I-MP framework is illustrated in Fig. \ref{fig:1}. I-MP comprises three modules: environment understanding (EU), reactive planner (RP), and low-level controller (LC). EU extracts multimodal environmental interaction features by first topologizing complex environments (Section \ref{sec_3_1}), and then build energy-based contact model (Section \ref{sec_3_2}). RP incorporates interaction features as computational constraints, leveraging fixed-point theory to determine the robot's convergence domain $\mathcal{X}_{con}$. Then, the Hausdorff distance discriminate plausible states $x^* \in \mathcal{X}_{con}$ to the motion intents $ \in \mathbb{R}^3$. Motion intent serves as the transition parameter from the perception-motion coordination mechanism \cite{de2022insect, yoshioka2024sensory, rao2024sensory, joshi2023dynamic}, aiming to resolve conflicts between kinesthetic perception \cite{sintov2023simple} and motion tasks. Subsequently, RP generates motion state variations ($F$ or $\Delta v$) by solving the inverse problem of shifting from current to planned states (Section \ref{sec_3_3}). LC executes these variations as actuation commands for task completion.

We deploy the I-MP framework on a robotic arm system equipped with full-body e-skin, integrating proximity with force sensing capabilities \cite{zhou2023tacsuit}, and construct simulation and hardware test platforms for stress testing and baseline comparison. The simulation environment is set with densely- and randomly-arranged fixed and movable rigid objects. We further incorporate objects with elasticity and plasticity into the hardware tests. The experimental results show the desirable motion adaptation of robots in cluttered environments with I-MP, and the robot can actively interact with objects while ensuring the continuity of relative velocity between itself and objects, thereby avoiding collisions. 
\subsection{Outline}
\label{sec_1_3}
In Section \ref{sec_2}, we formulate the interactive motion planning problem along with the relevant notation. The proposed method is outlined in Section \ref{sec_3}. The experimental platform is described in Section \ref{sec_4}. In Section \ref{sec_5}, the performance of the I-MP method is systematically evaluated. Section \ref{sec_6} is dedicated to validating the framework s reliability in real-world scenarios, including a free-motion space expansion task.
\section{Problem statement and notation}
\label{sec_2}
\subsection{Interactive Motion Planning Problem}
\label{sec_2_1}
Consider a robotic workspace $\mathcal{W} \in \mathbb{R}^3$ containing numerous objects $\mathcal{O}(t) \in \mathbb{R}^3$, each possessing its own environmental features $\mathcal{M}:=\{W, \theta\}$, where  $W$ is the geometry features and $\theta$ is the interaction features. The robot is modeled as a discrete control system  $\mathcal{X}_{r}(t) \in \mathbb{R}^3$. The goal of interactive motion planning is to determine the convergence-conformant region $\mathcal{X}_{con}(t) \in \mathbb{R}^3$ the set of reachable states at time $t+1$ based on the robot's spatial position and the environmental interaction features $\theta$ at time $t$. From this region, an optimal spatial target $x^*$  is planned for tracking based on target domain $\mathcal{G}_g \in \mathbb{R}^3$. The planner then computes the motion variations ($F$ or $\Delta v$) required to move from the current state to the planned state, based on the interaction features $\theta$. We denote this interactive motion planning problem as: $\mathcal{P} = \{\mathcal{X}_{r}(t), \mathcal{G}_g, \mathcal{M}\}$.

\subsection{Autonomous Contact Modeling Problem}
\label{sec_2_1}
Consider an object $\mathcal{O}$ with unknown interaction features $\theta$ located at position $x_{o, t}$ within the workspace. The robot applies a sequence of interactive forces $F(0:t)$ to the object and acquires time-series stimuli-responses pair $\psi(0:t)$ at the interaction interface. These interaction data are subsequently used to estimate the interaction features $\theta$ of object $\mathcal{O}$, such that a perception confidence constraint $\sigma(\theta, \theta^{-1})$ is satisfied. The autonomous contact modeling problem is denoted as: $\mathcal{C} = \{\mathcal{X}_{r}(t), x_{o, t}, \sigma\}$.

\section{Methodology}
\label{sec_3}
\subsection{Environment Understanding}
\label{sec_3_1}
The object of this subsection is to propose a method for perceiving environmental interaction features $\theta$ for online contact modeling. 
We model the robot system as a second-order dynamic system with $n$ degrees of freedom, whose pose is denoted as $\bar{x}(t)=(x(t), \Psi\!(t))\in\mathbb{R}^2\times\!S^n$, and the spatial state and actuation input are written as $\mathcal{X}_{r}(t):=(\bar{x}(t), (r, h))$ and $\tau\!(t)\in\!S^n$, respectively.

In this study, the workspace is assumed to be an abstract 2D flat world $\mathcal{W}(t, x):=\{\mathcal{G}(t),\mathcal{O}(t), \mathcal{X}(t), \mathcal{F}(t), \varepsilon(t)\}$ encompassing the target domain $\mathcal{G}(t)$, object domain $\mathcal{O}(t)$, robot domain $\mathcal{X}(t)$, local energy domain $\varepsilon(t)$, and free-motion domain $\mathcal{F}(t)$.

The target domain, a circular area of fixed size, is denoted as:
\begin{equation}
    \mathcal{G}(t):=\{x\in\mathcal{W}| |x-g|\leq\!r_g\},
\end{equation}
where $g\in\mathbb{R}^2$ is the target point and $r_g$ is the target domain radius, and functions to constrain the position of the robot that converges to the target point. 

The occupied domain is determined by the geometric outline of objects that contain unknown objects $O_k$, operable objects $O_m$, and inoperable objects $O_f$, which are denoted as the uncertain domain $\mathcal{O}_k(t)$, operable domain $\mathcal{O}_m(t)$, and inoperable domain $\mathcal{O}_f(t)$, respectively. All objects, defined by plane geometry and augmented by the Z-axis direction, are signified as $O=\{O_1, ..., O_m\}$. The occupied domain can then be written as:
\begin{equation}
    \label{equa_1}
    \mathcal{O}(t):=\{x \in \mathcal{W} \mid \cup \, O_{m/f, i}\}.
\end{equation}

It is noted that the robot merely takes an interest in any implicit physical properties of objects that block its movement.

The local energy domain is a fixed-size circular region corresponding to the robot's perception domain. The intradomain environment and objects are represented by the parameter of energy, which can be denoted as:
\begin{equation}
    \varepsilon(x):\{x\in\mathcal{W} ||x-x(t)| \leq r_p\},
\end{equation}
where $r_p$ represents the sensing range of the proximity sensor. 

The free-motion domain is marked as the difference set between the robot domain and the occupied domain inside the parameter of world $\mathcal{W}$, which can be expressed as:
\begin{equation}
    \mathcal{F}:=\{x\in\mathcal{W}\mid C_{\mathcal{W}}\mathcal{X}_r(t) \setminus \mathcal{O}(t)\}.
\end{equation}

After topologizing the geometric properties of the objects, we formulate the extraction problem of the implicit physical property as a parameter estimation, and define such parameters as operable vectors $\theta = [K, D, C]^T$, by incorporating the spring constant ($K$), damping coefficient ($D$), and displacement constant ($C$). The perceptual modality of the object $O_m$ is presented as $\psi(t)$:

\begin{equation}
    \label{equa_2}
    \psi(t)=[\Delta\!x_m, \dot{x}\,, F_m]^T,
\end{equation}
where $\Delta\!x_{m}, \dot{x}\,, f_m $  refers to the interaction displacement, velocity, and force with the object $O_m$ in the world frame $W$.

Specifically, the interaction displacement can be obtained by calculating the position difference of the contact point $p_{m}$ over the time series:
\begin{equation}
    \Delta\!x_{m, t} = p_{m, t}-p_{m, t-1},
\end{equation}
where the mapping of contact point ${}^{W}p_{Z}=[p_{m}, 1]^T \in \mathbb{R}^4$ with object $ \mathcal{O}_{m}$ in the word frame $W$, can be described as the coupling between the position mapping  ${}^{J}p_{Z} \in \mathbb{R}^4$ from sensing sensor frame $Z$ to adjacent joint frame $J$ and the pose mapping ${}^{J}_{W}T \in SE(3)$ from adjacent joint frame $J$  to world frame $W$. $\Delta\!x_m(0:t)$ represent the collected contact displacement from the beginning of contact to time $t$.

The interaction velocity $ \dot{x}$ is denoted as:
\begin{equation}
    \dot{x}_{m, t} =J(q_{t})\dot{q}_{t},
\end{equation}
where $J \in \mathbb{R}^j$ is the Jacobian transpose from the adjacent $J$  joint space to the Cartesian space. $q \in \mathbb{R}^j$ and $\dot{q} \in \mathbb{R}^j$ are the joint angle and joint velocity, respectively. $\dot{x}_m(0:t)$ represent the collected interaction velocity from the beginning of contact to time $t$.

The contact force can be observed by covering a large area of the robot's body with tactile sensors. The interaction force $F_m$ is denoted as:

\begin{equation}
    F_{m, t}=\prescript{J}{W}{R}F_{z, t},
\end{equation}
where $F_{m, t}$ is the observed interactive force at contact point,  ${}^{J}_{W}R \in SE(3)$ is the rotation transformation matrix from sensing sensor frame $Z$  to world frame $W$. $F_m(0:t)$ represent the collected interaction force from the beginning of contact to time $t$.

We model the operation parameter inference into a data-driven system identification problem \cite{ding2023least}, and presume the linear projection from the observed data to the parameter vector $\theta$:
\begin{equation}
    \boldmath{Y}(t)=\boldmath{H}(t)\theta+\boldmath{V}(t),
\end{equation}
where $\boldmath{Y}(t)=[y(0), y(1), \dots, y(t)]^T \in \mathbb{R}^{t}$ is stacked output vector, $\boldmath{H}(t)=[\psi(0), \psi(1), \dots, \psi(t)]^T \in \mathbb{R}^{t \times 3}$ is stacked information matrix, and $\boldmath{V}(t)=[v(0), v(1), \dots, v(t)]^T \in \mathbb{R}^{t}$ is a stacked noise vector with mean zero and variance $\sigma^2$.

Values of these parameters can be determined using the least squares (LS) parameter estimation \cite{ding2014combined, distefano2014comparison}:

\begin{align}
    J_1(\theta) &:= \sum_{j=1}^{t}v^2(j) = \sum_{j=1}^{t} [y(j) - \psi^T(j) \theta]^2 \nonumber \\
                &= \boldsymbol{V}^T(t) \boldsymbol{V}(t) \nonumber = (\boldsymbol{Y}(t) - \boldsymbol{V}^H(t) \theta)^T (\boldsymbol{Y}(t) - \boldsymbol{V}^H(t) \theta) \nonumber \\
                &= \left\lVert \boldsymbol{Y}(t) - \boldsymbol{V}^T(t) \theta \right\rVert^2.
\end{align}

We assume that the minimum value of $J(\theta)$ is reached when $\theta=\hat{\theta}$. If the partial derivative of $J(\theta)$ with respect to $\theta$ is set at zero, then:
\begin{equation}
    \label{equa_j_theta}
    {\frac{\partial J_1(\theta)}{\partial \theta}} \Big|_{\theta=\hat{\theta}} = -2\boldsymbol{H}_{t}^{T}(\boldsymbol{Y}_t - \boldsymbol{H}_t\theta) \Big|_{\theta=\hat{\theta}} = 0.
\end{equation}

The equation \ref{equa_j_theta} can be equivalent to:
\begin{equation}
    \label{equa_h_theta}
    (\boldsymbol{H}^T_t \boldsymbol{H}_t) \hat{\theta}_t = \boldsymbol{H}^T_t \boldsymbol{Y}_t.
\end{equation}

The matrix $ (\boldsymbol{H}^T_t \boldsymbol{H}_t)$ in equation \ref{equa_h_theta} is a positive definate matrix for tactile perception pair $\psi(t)$ of time-series observation. The interactive features $\theta$ can be estimated by the LS estimate:

\begin{equation}
    \hat{\theta}_t =  (\boldsymbol{H}^T_t \boldsymbol{H}_t)^{-1} \boldsymbol{H}^T_t \boldsymbol{Y}_t.
\end{equation}

A more practical expression of the interactive feature parameter estimation is:

\begin{equation}
    \hat{\theta}_t =  \left[\sum_{j=1}^{t} \psi(j) \psi^{T}(j)\right]^{-1} \left[\sum_{j=1}^{t}\psi(j) y(j) \right].
\end{equation}

Considering the random observation noise of robot joints and e-skin sensors, the goal of LS estimation is to converge to the true value of the interaction features as the number of tactile perception pairs increases. We formulate the criterion function for the LS estimation of the interaction features as:

\begin{align}
J_1(\hat{\theta}_t) 
:= & \left[\boldsymbol{Y}_t - \boldsymbol{H}_t \hat{\theta}_t \right]^T \left[\boldsymbol{Y}_t - \boldsymbol{H}_t \hat{\theta}_t \right] \nonumber\\
= & \left[ \boldsymbol{Y}_t - \boldsymbol{H}_t(\boldsymbol{H}^T_t\boldsymbol{H}_t)^{-1} \boldsymbol{H}^T_t\boldsymbol{Y}_t \right]^T \nonumber\\
  & \cdot \left[ \boldsymbol{Y}_t - \boldsymbol{H}_t(\boldsymbol{H}^T_t\boldsymbol{H}_t)^{-1} \boldsymbol{H}^T_t\boldsymbol{Y}_t \right] \nonumber\\
= & \boldsymbol{Y}_t^T \left[\boldsymbol{I}_t - \boldsymbol{H}_t(\boldsymbol{H}^T_t\boldsymbol{H}_t)^{-1} \boldsymbol{H}^T_t\right]^T \nonumber\\
  & \cdot \left[\boldsymbol{I}_t - \boldsymbol{H}_t(\boldsymbol{H}^T_t\boldsymbol{H}_t)^{-1} \boldsymbol{H}^T_t \right]\boldsymbol{Y}_t \nonumber\\
= & \boldsymbol{Y}_t^T \left[\boldsymbol{I}_t - \boldsymbol{H}_t(\boldsymbol{H}^T_t\boldsymbol{H}_t)^{-1} \boldsymbol{H}^T_t\right]^2 \boldsymbol{Y}_t \nonumber\\
= & \boldsymbol{Y}_t^T \left[\boldsymbol{I}_t - \boldsymbol{H}_t(\boldsymbol{H}^T_t\boldsymbol{H}_t)^{-1} \boldsymbol{H}^T_t\right]\boldsymbol{Y}_t,
\label{eq:final}
\end{align}
where $\boldsymbol{I}_t$ is the identity matrix with the size of $t \times t$.

Encoding the operational parameters $\hat{\theta}$ into work metrics can help robots construct an energy-cost model for physical interactions.

\subsection{Contact Modeling and Environment Representation}
\label{sec_3_2}
To alleviate the computational burden introduced by high-dimensional interaction models, this study uniformly represents environmental interaction features in the form of work \cite{khatib1986real, tzes2022reactive, arslan2019sensor} , enabling their linear superposition in the workspace. The representations are bounded within the local energy domain to mitigate the calculation load when motion planning requirements are met.

In this study, the target point is designed as an attractive potential field, $\mathbf{U}_{g^{*}}(x)$ whose linear, elastic features drive the robot to planned states. 
\begin{equation}
    \mathbf{U}_{g^{*}}(x)=\frac{1}{2}k_g(x^{*} - x(t))^2,
\end{equation}
where $k_g$ is a hyperparameter of $\mathbf{U}_{g^{*}}(x)$, designed manually.

The viscosity field $\mathbf{U}_{f}(x)$ is employed as well to represent the free-motion domain that considers the maximum velocity of the robot. The damping potential energy difference from the current state to the target state can be expressed as:
\begin{equation}
    \mathbf{U}_{f}(x)=\frac{1}{2}k_d(x^{*} - x(t))\dot{x}(t),
\end{equation}
where $k_d$ is a hyperparameter of $\mathbf{U}_{g^{*}}(x)$, designed manually. $k_d$ and $k_g$ are jointly used to regulate the robot's cruising speed and motion rigidity.

An artificial viscosity field is included to confine the velocity within the safe range. $W=D_o{\dot{x}^2}$ expounds the virtual power loss.

In terms of operable objects, the operational energy cost $O_m$ can describe the inner product of the operational vectors $\theta$ and the spatial state of the object.
\begin{equation}
    \mathbf{U}_{o, m}(x)=\sum\theta_i\psi_{m,i}.
\end{equation}

The operational vectors of unknown objects are assumed to follow a linear elastic model facilitating kinesthetic perception. The energy cost can be represented as:
\begin{equation}
    \mathbf{U}_{n, m}(x)=\Delta\!x_m \, K_M.
\end{equation}

$\mathbf{U}_{n, m}$ will drive robots to apply interactive forces $F_{act}$ to unknown objects. $F_{safe}$ is the maximum force that robot can interact safely with the objects, The interaction force pulse can be determined as:

\begin{equation}
    F_{act} = F_{safe}sin(\omega t), t \in (0, 100],
\end{equation}
where $\omega$ is the impulse frequency, $t$ is the number of interactions which is coordinate with the number of the observation.

To avoid undesired collisions and enable the robot's interactions with objects, we introduce a dedicated viscous field $\mathbf{U}_{o,v}(x)$ to represent the restricted area around objects with unknown properties. Ideally, the robot's velocity towards the object should converge to zero to ensure that during physical contact, the robot continuously shifts from a dynamic state to a static state. In practice, a safe speed $V_{safe}=1mm/s$ has been added to the boundary conditions to simplify the solving process for the repulsive potential coefficient of the object.The robot's relative position to the objec $O_i$ can be denoted as:
\begin{equation}
    \label{equa_4}
    d_i = o_i^* - x^*
    (o_i^*, x^*) \in \underset{o_i\in O_i, x \in \mathcal{X}_r(t)}{arg \quad \mathrm{min}}{d(o_i, x_i)}.
\end{equation}

The viscous field needs to regulate the robot speed so that it converges to $v_{safe}$ before the robot contacts objects, and the equation of the robot motion state in the workspace is modeled as:
\begin{equation}
    \label{equa_5}
    \ddot{x} = M(q)^{-1}(F^* - C(q, \dot{q})\dot{x} - G(q))
\end{equation}

Considering the interaction with an unknown object $O_i$, the energy gradient is obtained:
\begin{equation}
    F^*=k_p(g^*-x(t))-(D_i+D_W)\dot{x}(t).
\end{equation}

The initial conditions are written as:
\begin{equation}
    x(0)=0
\end{equation}
\begin{equation}
    \dot{x}(0)={k_p(g^*-x(t))}/{D_w}.
\end{equation}

By solving the critical damping$D_W^*$, a safe robot traveling speed $\dot{x}(t)=v_{safe}$ during its contact with objects is acquired.

This dedicated viscous field would be converted into a repulsive potential field  $\mathbf{U}_{o, r}(x)$ to guide the robot away from inoperable objects. Detailedly, the viscous field surrounding a fixed object can be transformed into a repulsive potential field that is artificial, linear, and elastic, with robots being kept away from inoperable objects. At the repulsive field center sits the maximum distance point from the set $O_i\cap\varepsilon(t)$ to the set $B=\mathcal{X}(t)\cup\mathcal{G}(t)$, which is based on the Hausdorff distance to avoid saddle points in the energy field.
\begin{equation}
    \mathbf{U}_{o, r}(x) = \frac{1}{2} k_0 (o^* - x(t))^2.
\end{equation}

\begin{equation}
    h(O_i, B) = \max_{o^* \in O_i \cap \varepsilon(t)}.
            \Bigl\{ \min_{b \in B} \, d(o^*, b) \Bigr\}.
\end{equation}

Therefore, the energy state at the robot's position $x(t)$ can be written as:
\begin{equation}
    \mathbf{U}(x(t)) =\mathbf{U}_{g^*}(x(t)) + \mathbf{U}_{f}(x(t)) + \sum_{i} \mathbf{U}_{o}(x(t)).
\end{equation}

\begin{figure*}[!t]
\centering
\includegraphics[width=\textwidth]{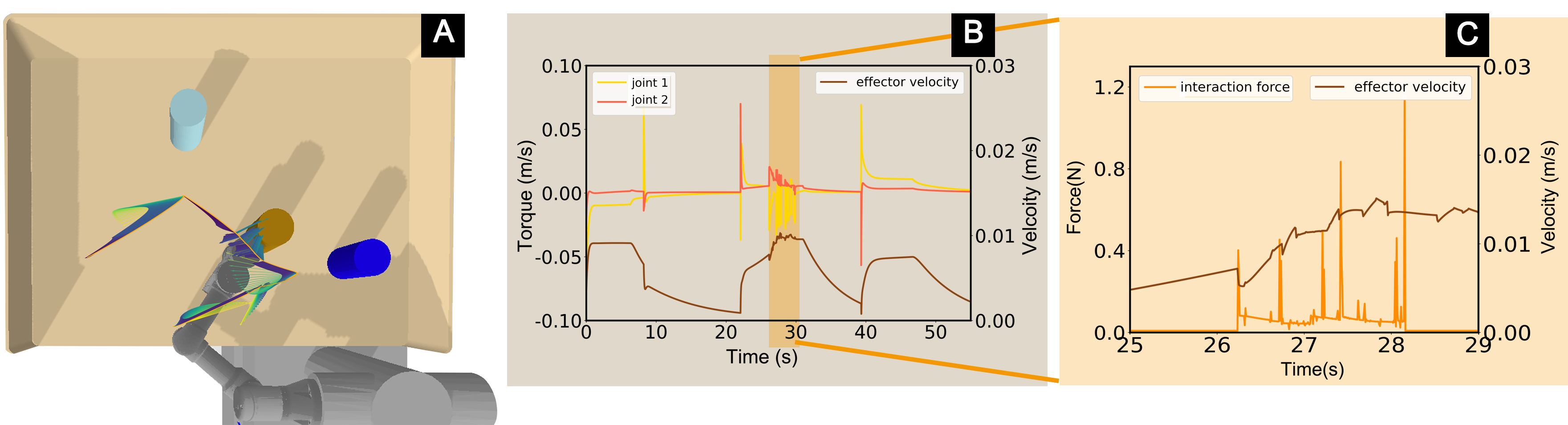} 
\caption{Simulated robot interactions with operable objects. (A): The trajectory and planned motion variables of the robot that traverses three target points. (B): Joint torques and end-effector velocity of the robot. (C): Interaction forces and end-effector velocity during robot interaction with objects.}
\label{fig:2}
\end{figure*}

\begin{figure}[!t] 
\centering 
\includegraphics[width=3.5in]{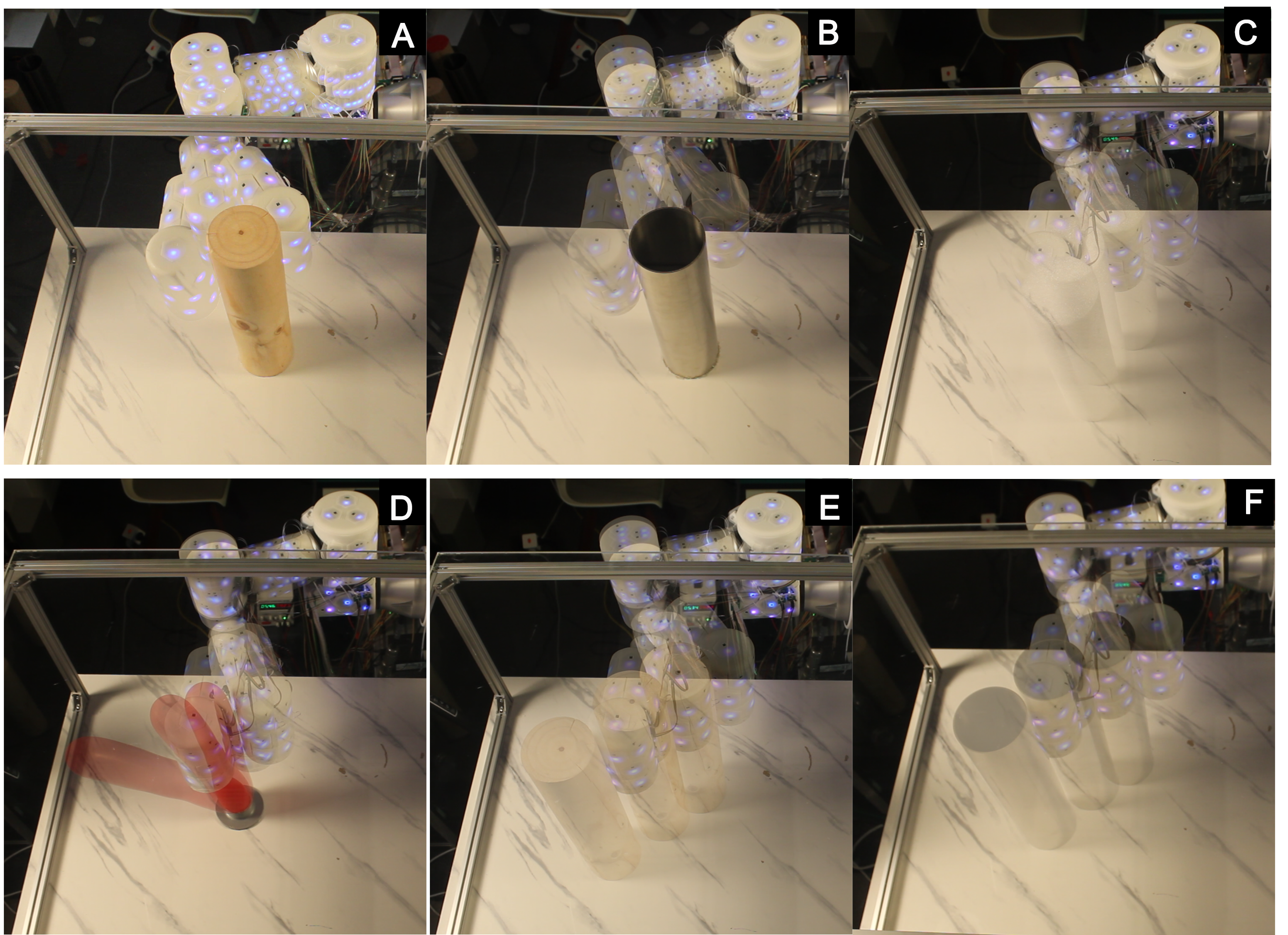} 
\caption{Hardware tests where the robot interacts with varying object types: fixed wood (A), fixed steel (B), movable foam (C), fixed elastic balloons (D), movable wood (E), movable steel (F).}
\label{fig:3} 
\end{figure}

\subsection{Interactive Motion Planner}
\label{sec_3_3}
Two functionalities, namely target point approaching and kinesthetic perception \cite{sintov2023simple} are found in the interactive motion planner. The conflicts between both functionalities are resolved by motion intents identified by perception confidence. The planner takes motion intents and interactive features as computational constraints to discriminate plausible special states. By solving the inverse problems of current-to-planned states, the planner drives the robot to perform motion tasks.

We define the kinesthetic perception tasks as $s_a$ and target point approaching as $s_p$. Perception confidence aims to resolve the binary classification problem $s=[s_a, s_p]$ based on the perception error of object contacts. The generated motion intents serve as constraints for the RP in planning states $x^*$.

The planned states $x^* = [g^*, k^*]$  are confined by the motion intents $s$ and environmental features $\mathcal{M}(t):=\{W, \theta\}$, where $g^*$ also stands for the mapping of the target point $g$ at the local energy domain boundary $\varepsilon(x)$ supported by the minimum distance principle, which can be written as:

\begin{equation}
\varphi: g \rightarrow \varepsilon(x)
\end{equation}

\begin{equation}
g^* = 
\begin{cases}
\underset{x \in \varepsilon(x)}{\arg\min} \, d(x, g) & \text{if } g \notin \varepsilon(x)\\[1.2ex]
g & \text{if } g \in \varepsilon(x)
\end{cases}
\end{equation}

The planned state of kinesthetic perception $k^*$ maps objects $O_m$ at the boundary of the expected displacement domain $\mathcal{K}(x, \Delta d)$ as guided by the minimum distance principle:

\begin{equation}
\varphi: o_m \rightarrow \mathcal{K}(x, \Delta d)
\end{equation}

\begin{equation}
\Delta d = \sin(\omega t) \, \Delta d_{\max}, \quad t \in (0, 100]
\end{equation}

\begin{equation}
k^* = \underset{x \in \mathcal{K}(x, \Delta d)}{\arg\min} \, d(x, o_m)
\end{equation}
where $\Delta d_{max} = F_{safe}/K_m$ marks the maximum interactive displacement to confine the interactive forces $F_{act}$ within the safe threshold $F_{safe}$.

Subsequently, the actuation force $F$ in Cartesian space is obtained by solving the energy gradient that flows from the current to the planned energy state:
\begin{equation}
    \label{equa_9}
    F^* = -grad[\mathbf{U}(x, x^*)].
\end{equation}

A robotic system characterized by $n$ degrees of freedom can be identified as the impedance system or the admittance system to generate command vectors. The simple admittance system approach directly maps the actuation force into the joint space as the actuation input. The command vector $\tau^*$ in the joint space, which contributes to changes in the motion state, is illustrated followingly:
\begin{equation}
    \label{equa_10}
    M(q)\ddot{q} + C(q, \dot{q})\dot{q} + G(q) = \tau^* - \tau_{ext},
\end{equation}
where $\tau = J(q)^T F^*$ denotes the mapping of the energy gradient from Cartesian space to the joint space and $J(q) \in \mathbb{R}^{n \times 6}$ represents the Jacobian matrix, while $\tau_{ext} = J(q)^{-1} F^*$ marks the mapping of the interaction force from Cartesian space to the joint space. $M(q)$, $C(q, \dot{q})$, and $G(q)$ represent the represent the inertial and Coriolis matrices, and gravitation vector, respectively.

The rate of change in velocity ($\Delta v$) in Cartesian space can be denoted as follows:
\begin{equation}
    \ddot{x} = M(q)^{-1}(F + F_{ext} - C(q, \dot{q})\dot{x} - G(q) + \omega),
\end{equation}
\begin{equation}
    \label{equa_delta_v}
    \Delta v = \int \ddot{x} dt,
\end{equation}
where $\dot{x}$ and $q$ stand for the speed vector of the robot in the workspace and the joint angle vector in the joint space, respectively, and $\omega$ denotes the motor noise.

\subsection{Low Level Control}
The objective of the LC is to map the motion state variations planned by the RP—namely, the force $F^*$ (Equation \ref{equa_9}) or the velocity increment $\Delta v$ (Equation \ref{equa_delta_v})—from Cartesian space to joint space, thereby driving the robot to execute motion commands.

The Jacobian matrix $J$ maps the relative motion in Cartesian space to joint space:
\begin{equation}
\begin{aligned}
\tau_{\text{ext}} &= J(q)^{-1} F^*, \\
\dot{x} &= J(q)\dot{q}, \\
\ddot{x} &= J(q)\ddot{q} + \dot{J}(q)\dot{q}, \\
\ddot{q} &= J(q)^{-1}(\ddot{x} - \dot{J}(q)\dot{q}).
\end{aligned}
\end{equation}

In impedance control mode, the objective of the LC is to map the driving force $F^*$ planned by the RP to joint space and compensate for the system dynamics:

\begin{equation}
    M(q)\ddot{q} + C(q, \dot{q})\dot{q} + G(q) + \tau_{ext} = \tau^*.
\end{equation}

The joint motors achieve safe and compliant free motion space expansion by tracking the joint driving torque $\tau^*$ computed by the LC in real time.

In admittance control mode, the LC aims to map the rate of change in velocity ($\Delta v$) planned by the RP to joint space to achieve real-time velocity control  $\dot{q}^*$:
\begin{equation}
    J(q)^{-1}(\Delta v + \dot{J}(q)\dot{q}) = \dot{q}^*,
\end{equation}
where $\dot{J}(q)$ is the derivative of the Jacobian $J(q)$.

\section{Experimental Platform}
\label{sec_4}
We developed a wheeled humanoid robot system both in the PyBullet simulator \cite{coumans2016pybullet} and in real-world experiments. As shown in Fig. \ref{fig:1}A, it contains two 6DoF (six degrees of freedom) robotic arms supported by a whole-body multi-modal tactile sensing system that empowers the robot to implement contact-rich upper-limb tasks. We made this electronic skin (e-skin) based on our previously introduced TacSuit \cite{zhou2023tacsuit} sensors. The controller processes and maps sensor data to the Cartesian space through kinematic chains by referring to the configuration space representation.

The simulation setup consists of a 1.2$\times$0.9m tabletop randomly occupied by rigid cylindrical objects, each of which has a 10cm diameter. Each object, with a friction coefficient set at 0.5 and a uniformly distributed mass, was randomly assigned as either fixed or movable for testing. Furthermore, we have augmented the hardware tests introducing objects with diverse mechanical properties, such as steel, wood, foam, and balloons. These materials exhibit significant differences in stiffness and friction, features that are challenging to simulate with high fidelity.

\section{Statistical Performance of I-MP}
\label{sec_5}
This chapter presents the excellent performance of I-MP in autonomously expanding the free motion space through both simulation and physical experiments. We designed three sets of experiments—motion-control continuity test, baseline comparison test, and stress test—to systematically evaluate I-MP's capability in addressing the unsolvable free-motion space and achieving safe interaction.
\subsection{Motion-Control Continuity}
Motion-control continuity refers to the smooth transformation of velocity as robots interact with objects \cite{odhner2014compliant, schumacher2019introductory}. The dramatic changes in velocity resulting from the rapid dissipation of energy reveal the potential physical dangers \cite{skrinjar2018review, gilardi2002literature}. Meanwhile, motion-control continuity can function as a valuable metric of the interactive performance of planners who should regulate the relative velocity between the self-entity and objects to ensure safety as robots move in cluttered environments.

The robot's joint space actuation, end-effector velocity, and interaction forces during task execution are analyzed. Both the interaction force and end-effector velocity remained stable with no occurrence of catastrophic collisions (see Fig. \ref{fig:2}, \ref{fig:3}) throughout the robot interactions with a variety of object types (see MovieS1, Supplementary Fig.1). As shown in Fig. \ref{fig:2}B, the robot's velocity remained stable within the range of 0-0.01 $m/s$ throughout the task execution, exhibiting smooth and continuous variations. During interaction with objects, the contact force was consistently maintained between 0 and 1.2 $N$, with an initial contact force of only 0.4 $N$ (See Fig. \ref{fig:2}C). These results demonstrate that I-MP can reliably ensure the safety of robot-environment interaction by managing motion-control continuity. 

\begin{figure}[!t] 
\centering 
\includegraphics[width=3.5in]{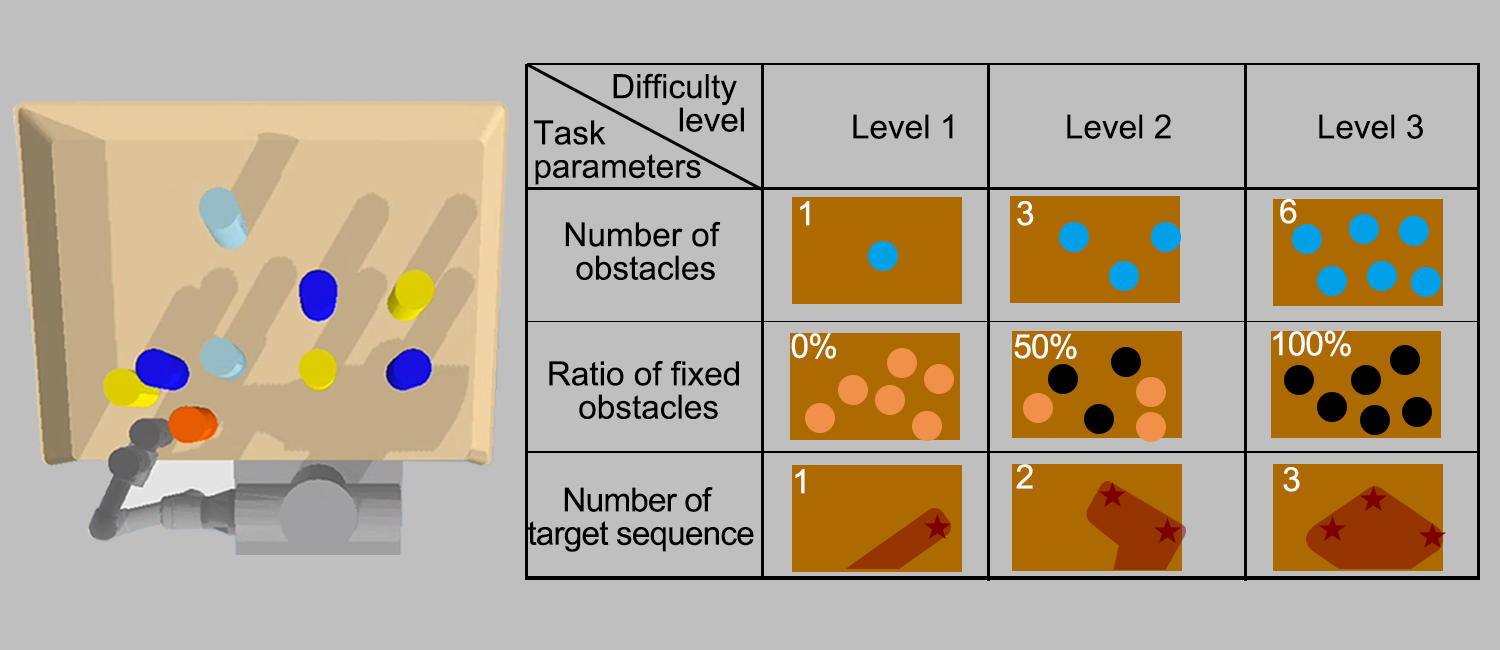} 
\caption{Design of test scenarios and elucidation of the task difficulty.}
\label{fig:4} 
\end{figure}

During physical interaction, the interaction velocity remained stable within the range of 0.1-0.7 $m/s$. The maximum contact force when in contact with non-manipulable objects was below 30 $N$, which corresponds to the upper limit of kinesthetic contact force. In contrast, contact forces with manipulable objects ranged between 0 and 3.5 $N$, reflecting safe interactive motion features (Supplementary Fig. S2 A).

From a temporal perspective, both the contact force and the relative motion velocity between the robot and various object types maintained strong continuity, consistent with the trends observed in simulation results.
\begin{figure}[!t] 
\centering 
\includegraphics[width=3.5in]{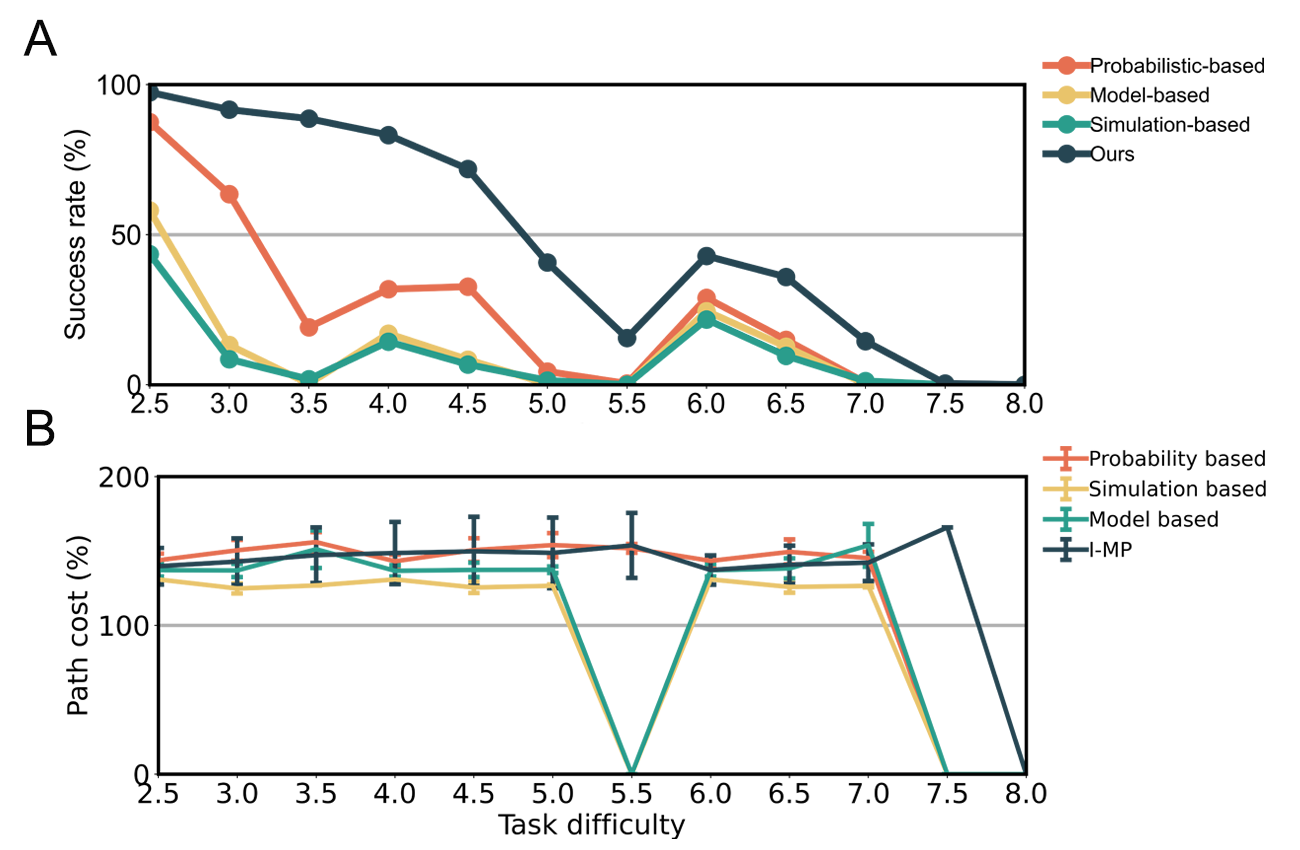} 
\caption{Baseline comparison of I-MP in success rate and path cost. (A) Comparison of success rates between I-MP and probability-, model-, and simulation-based planning methods. (B) Statistical differences between the I-MP-driven robotic motion paths and baseline method paths.}
\label{fig:5} 
\end{figure}

According to the simulation experiment on joint velocity and end-effector velocity of the robot in Fig. \ref{fig:3}, continuous variations upon contact with objects were maintained (see MovieS2). Additional experimental data comparing contact forces and motion velocities between simulation and hardware robot-object interactions are presented in Supplementary Fig. 2. The energy-based representation of the robot's Euclidean distance relative to unknown objects enables the robot to promptly adjust its velocity relative to the objects, thereby avoiding impacts resulting from the significant variation of velocity. Furthermore, the I-MP leverages implicit physical properties of objects inferred through kinesthetic haptic methods to avoid destructive interaction forces.

\begin{figure}[!t] 
\centering 
\includegraphics[width=2.5in]{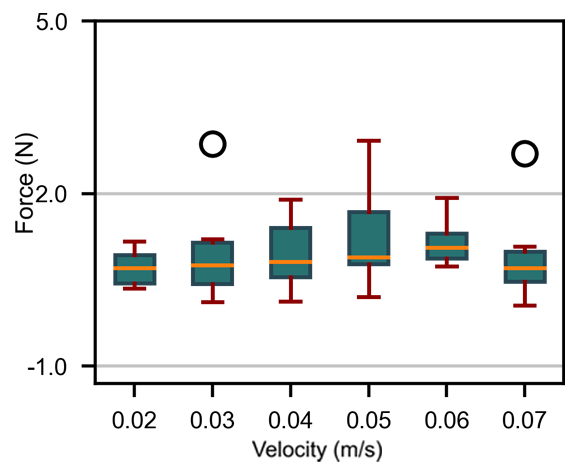} 
\caption{Statistical analysis of contact forces amid robot interactions with objects at cruising speeds ranging from 0.01 to 0.07 m/s.}
\label{fig:6} 
\end{figure}
We have employed a quantitative weighted approach to assess the experimental difficulty, which is written as $w=\sum\!a_in_i$(see Fig. \ref{fig:4}). The number of objects ($n_1$) is taken as the first difficulty factor, with the three provided conditions of 1 object, 3 objects, and 6 objects occupying 2.9$\%$, 8.7$\%$ and 17.4$\%$ of the space, respectively. The proportion of fixed objects is denoted as $n_2$  and ranks three levels: 0$\%$, 50$\%$, and 100$\%$. Task difficulty ($n_3$) is defined as the area ratio of the motion path to the testing space, scoring 15$\%$, 35$\%$, and 65$\%$ as the normal difficulty (one target point), moderate difficulty (two target points), and complicated tasks (three target points), accordingly.

\begin{figure}[!t] 
\centering 
\includegraphics[width=2.5in]{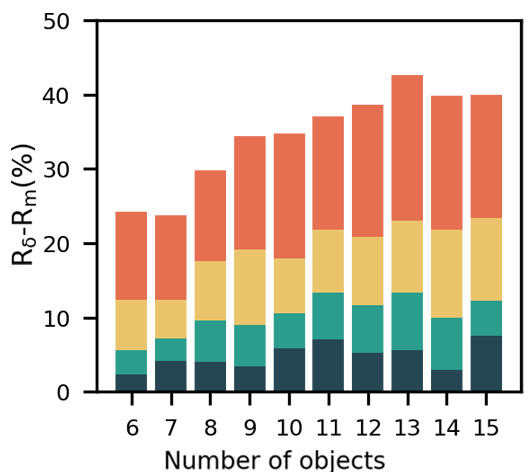} 
\caption{I-MP's success ceiling gap.}
\label{fig:7} 
\end{figure}

\subsection{Baseline Comparison}
We conduct comparative evaluations of I-MP using state-of-the-art approaches in physical hardware experiments and simulations (see MovieS3).

Probability-based, simulation-based, and model-based methods, as three promising planning approaches, are selected as baselines. Probability-based motion planning employs sampling-based algorithms to explore environments via random sampling, aiming to identify paths with maximum connectivity probabilities. We have also performed compliant control to track generated trajectories, ensuring safe interaction with objects and preventing catastrophic collisions. The simulation-based method utilizes a B-spline planning algorithm to generate reference trajectories, which are then evaluated in an external simulator (PyBullet) containing potential collision objects to verify execution feasibility and target reachability. Finally, we take the artificial potential field method as the model-based method. The experimental difficulty is determined by the number of objects, the fixed ratio of objects, and the path convergence rate.

\begin{figure*}[!t] 
\centering 
\includegraphics[width= 0.8 \textwidth]{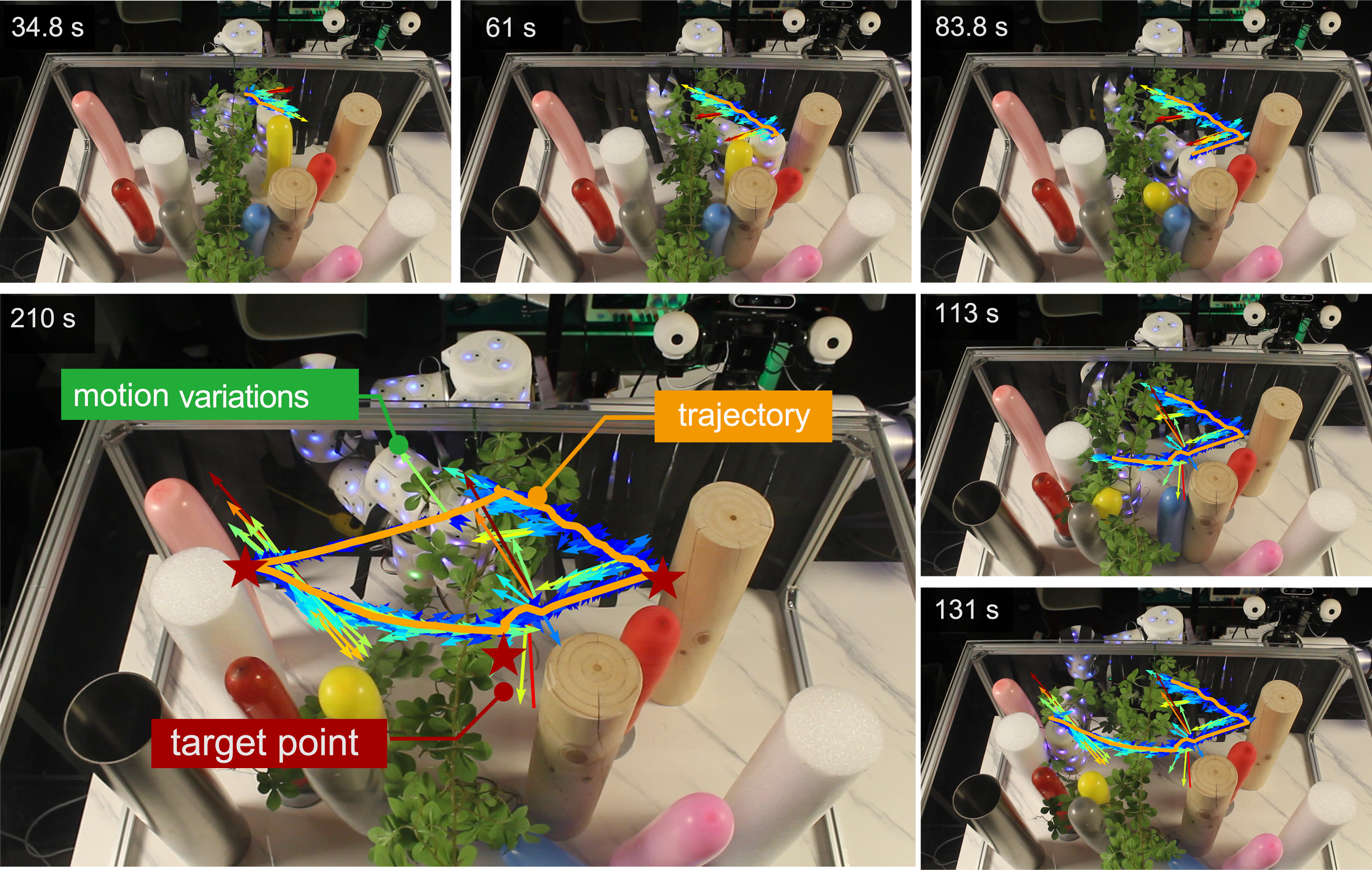} 
\caption{Clutter-based Motion Adaptation. The robot manages to identify operable or inoperable objects with its multi-modal e-skin, thereby reaching the target object by actively enlarging the motion space.}
\label{fig:8}
\end{figure*}

\begin{figure}[!t] 
\centering 
\includegraphics[width=2.5in]{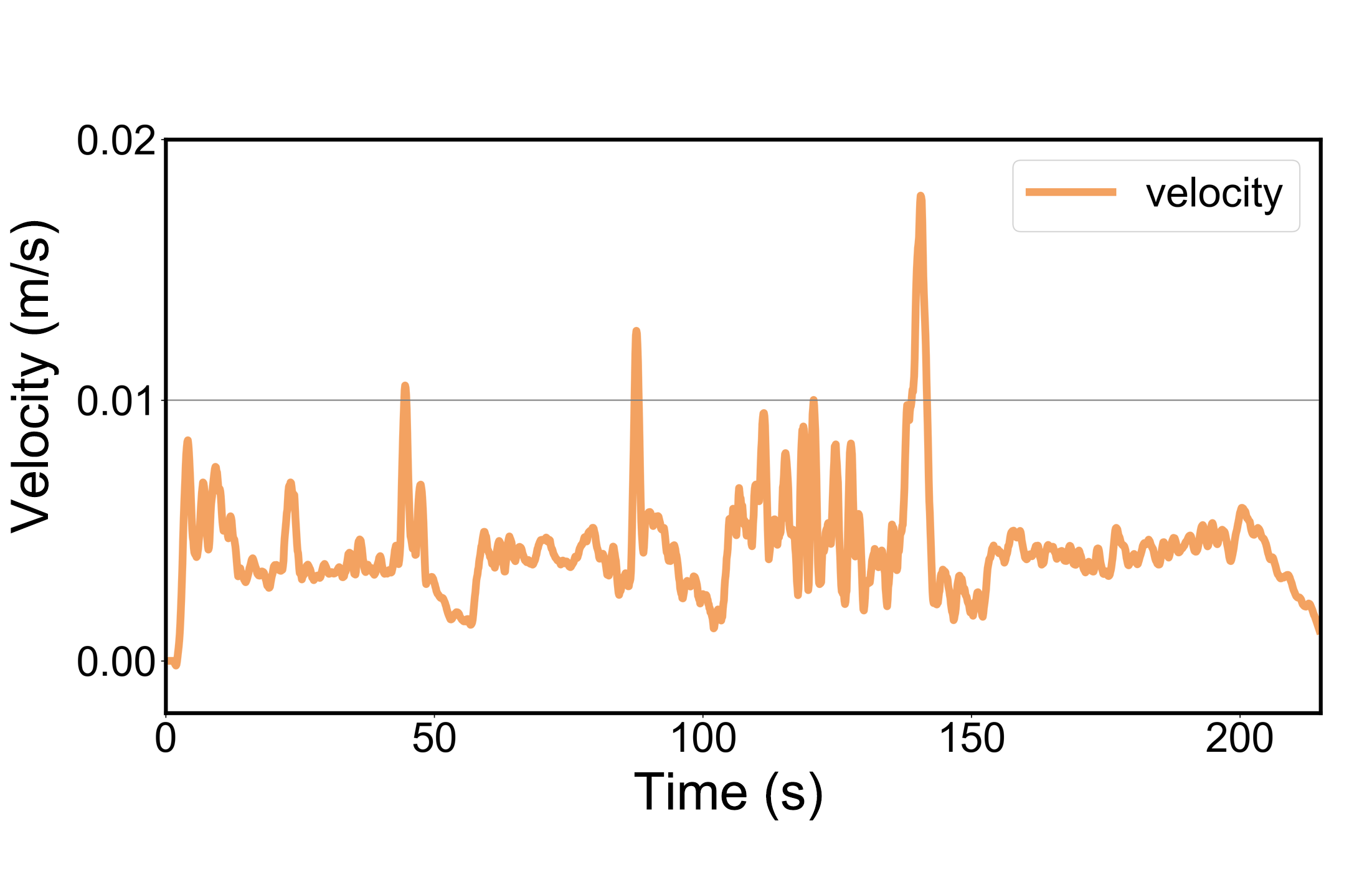} 
\caption{The velocity of the end effector during environmental exploration execution.}
\label{fig:9} 
\end{figure}

The testing workflow applied the randomly generated scenarios to evaluate four algorithms to eliminate environmental interference. Each experimental setting underwent 400 random simulations and 2 hardware experiments to ensure statistical reliability, resulting in a total of 43,200 simulations and 216 hardware experiments. A trial is deemed a failure if the contact force exceeds the 30-N threshold, or the end effector fails to reach the target positions. Task success rate and path cost make up of key evaluation metrics. In addition, we simplified the knock down problem by ensuring that the resultant force acting on the objects remains within a stability threshold, thus maintaining stability.

We compare the success rates of the proposed I-MP algorithm with those of baseline methods first, as shown in Fig. \ref{fig:5}A, and notably observe that I-MP demonstrates better adaptability than baseline methods, particularly within the [3.0, 5.0] task difficulty range. I-MP reached maximum success rate improvements of 69.51$\%$, 88.34$\%$, and 86.82$\%$ against probability-based, model-based, and simulation-based approaches, respectively. However, all methods failed in testing when task difficulty exceeded 7.5, which was primarily caused by the presence of dense obstacles, where model-based collision-free trajectory planners struggled to design feasible solutions. Furthermore, object-to-object contacts built up multi-body systems, making the inverse problem of interaction models intractable for motion planners. Additionally, we compare path costs across tested algorithms (see Fig. \ref{fig:5}B), with benchmark results showcasing I-MP's path cost premiums of 19.3$\%$ (vs. simulation-based), 6.32$\%$ (vs. model-based), and -1.77$\%$    (vs. probability-based) methods as the reasonable trade-off for its higher success rate and stabler real-time performance. 

In an effort to further validate the robot's capability in forestalling undesirable interaction impacts with objects, we have conducted tests with varied cruising velocities of the end-effector ranging from 0.01 to 0.07 $m/s$ for interaction with fixed, rigid objects. Simulation results indicate that the robot can effectively control the contact force within a reasonable range (see Fig. \ref{fig:6}).

\subsection{Stress Testing}
We designed stress test scenarios for other experiments, working to measure the optimization success rate under more difficult conditions. To this end, we increased the number of objects to a range of [6-15], with fixed ratios of 10$\%$, 20$\%$, 40$\%$, and 80$\%$. 500 randomized trials were conducted in each test setting, totaling 20,000 simulations. We first identified the feasibility of the generated trials and tested the I-MP with validated trials. Building on the results of stress tests, we chamfered the object bases to evaluate the accidental toppling impact of objects on the I-MP performance. 

As depicted in Fig. \ref{fig:7}, the success ceiling gap (i.e., the discrepancy between achieved success rates $R_m$ and upper bounds $R_\delta$) exhibits sensitivity to the fixed object ratios but stays robust against variations in object number (6-15 objects). The gap remains narrow ($<$5$\%$) at lower object ratios (0.1, 0.2) as an indicator of near-optimal performance and manageable interference. The ratio increased to 0.4 and 0.8 is found to progressively widen the gap by 11.8$\%$ and 16.6$\%$, respectively. Notably, the object number shows minimal impact as the gap fluctuates by $<$5$\%$ across tests, highlighting the algorithm's scalability in cluttered environments.

\begin{figure}[!t] 
\centering 
\includegraphics[width=2.5in]{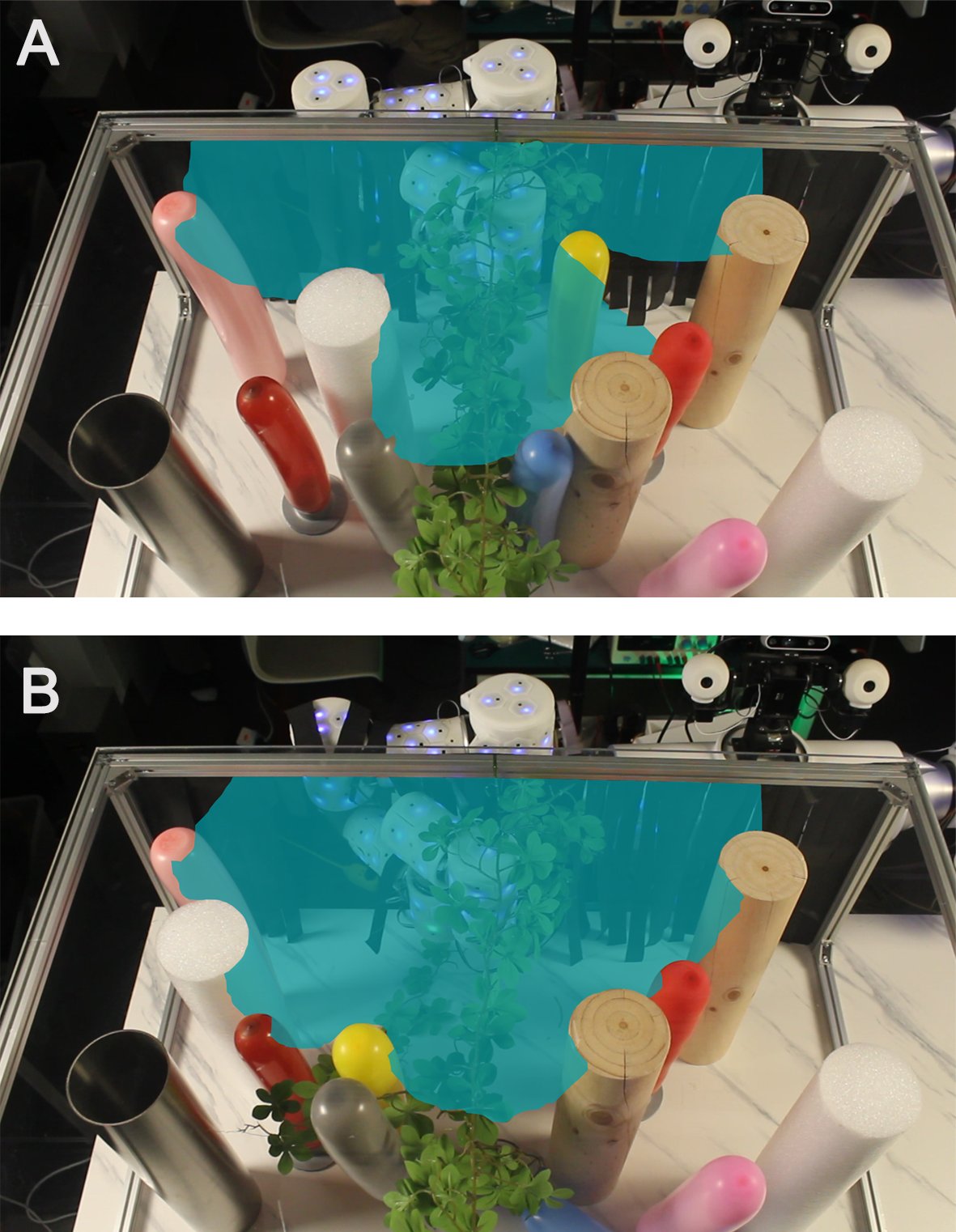} 
\caption{Free-motion space comparison before and after environmental exploration task.}
\label{fig:10} 
\end{figure}

\section{Real-World Experiments}
\label{sec_6}
After systematically evaluating the performance of I-MP, we further designed a more challenging scenario—the cabinet scenario—to validate its motion performance in complex environments that more closely resemble real-world applications. The interior of a cabinet is typically cluttered with various objects that occlude one another. When searching for a target item within it, humans must carefully maneuver and rummage through the clutter to overcome the challenges posed by the confined space and avoid hazardous collisions. \cite{doi:10.1177/0278364912471865, aucone2024synergistic, zhong2022soft}.

We instantiate the interaction as an environmental exploration task. The planner endeavors to maneuver the end effector of a robotic arm to various planned states by physically interacting with the surroundings. Specifically, when confronted with unsolvable free-motion space, the planner applies a sequence of force interactions on objects through robot arm control to observe force-displacement features at the interaction interface and infer operable vectors. On this basis, the robot can successfully traverse the planned states using flexible bodily movements.

Fig. \ref{fig:8} demonstrates an experimental trial example. Operable objects such as artificial vegetation, plastic balloons, and foam, as well as fixed objects like wood, are effectively identified. I-MP successfully drive the robot to complete the environmental exploration task (see MovieS4). The orange line and vectors in Fig. \ref{fig:8} represent the trajectory of the end effector and the motion variations, respectively. According to the trajectories, when driven by motion alterations, the robot could displace operable objects and navigate around inoperable objects. Throughout the interactions with different types of objects, the end-effector velocity remains stable, ranging from 0.0 to 0.02 $m/s$, with no occurrence of catastrophic collisions (Fig. \ref{fig:9}). 

Fig. \ref{fig:10} illustrates the pre-trial and post-trial motion space. After completing the environmental exploration task, the free motion space increased significantly by 37.5\%. This clearly demonstrates the effectiveness of I-MP in addressing the absence of feasible free motion solutions in cluttered environments.

\section{Conclusion}
\label{sec_7}
This paper proposes an interactive motion planning framework in which the planner enables the robot to actively engage in physical interaction with the environment, addressing the challenges of line-in-sight suffer and unresolvable free-motion space within cluttered environments. The I-MP achieves autonomous contact modeling through a perception-action loop, thereby ensuring the reliability of interactive action reasoning. Furthermore, to alleviate the computational burden of interactive planning and ensure real-time performance, we introduce a reactive motion planner. This planner employs fixed-point theorems to discriminate feasible spatial state, thus avoiding the computational cost associated with extrapolating on high-dimensional contact models.

Experimental results demonstrate that I-MP maintains smooth motion-control continuity when interacting with objects of various materials and dynamic properties, indicating its strong capability for safe interaction and scene generalization. In an environmental exploration task conducted in a cabinet scenario lacking feasible free-space solutions, I-MP successfully drove the robot to complete the exploration task, exhibiting effective expansion of the feasible free space and consistent motion-control continuity. These results validate the strong adaptability of I-MP in unknown and cluttered environments.

\bibliographystyle{IEEEtran}
\bibliography{references}

@article{
doi:10.1126/scirobotics.adf7843,
author = {Tobia Marcucci  and Mark Petersen  and David von Wrangel  and Russ Tedrake },
title = {Motion planning around obstacles with convex optimization},
journal = {Science Robotics},
volume = {8},
number = {84},
pages = {eadf7843},
year = {2023},
doi = {10.1126/scirobotics.adf7843}}

@article{
doi:10.1177/0278364912471865,
author = {Advait Jain and Marc D Killpack and Aaron Edsinger and Charles C Kemp},
title ={Reaching in clutter with whole-arm tactile sensing},

journal = {The International Journal of Robotics Research},
volume = {32},
number = {4},
pages = {458-482},
year = {2013},
doi = {10.1177/0278364912471865}}

@ARTICLE{
  8954627,
  author={Xia, Fei and Shen, William B. and Li, Chengshu and Kasimbeg, Priya and Tchapmi, Micael Edmond and Toshev, Alexander and Martín-Martín, Roberto and Savarese, Silvio},
  journal={IEEE Robotics and Automation Letters}, 
  title={Interactive Gibson Benchmark: A Benchmark for Interactive Navigation in Cluttered Environments}, 
  year={2020},
  volume={5},
  number={2},
  pages={713-720},
  doi={10.1109/LRA.2020.2965078}}

@article{
  aucone2024synergistic,
  title={Synergistic morphology and feedback control for traversal of unknown compliant obstacles with aerial robots},
  author={Aucone, Emanuele and Geckeler, Christian and Morra, Daniele and Pallottino, Lucia and Mintchev, Stefano},
  journal={Nature Communications},
  volume={15},
  number={1},
  pages={2646},
  year={2024},
  publisher={Nature Publishing Group UK London}
}

@article{khatib1986real,
  title={Real-time obstacle avoidance for manipulators and mobile robots},
  author={Khatib, Oussama},
  journal={The international journal of robotics research},
  volume={5},
  number={1},
  pages={90--98},
  year={1986},
  publisher={Sage Publications Sage CA: Thousand Oaks, CA}
}

@article{barraquand1991robot,
  title={Robot motion planning: A distributed representation approach},
  author={Barraquand, Jerome and Latombe, Jean-Claude},
  journal={The International journal of robotics research},
  volume={10},
  number={6},
  pages={628--649},
  year={1991},
  publisher={Sage Publications Sage CA: Thousand Oaks, CA}
}

@inproceedings{brouwer2024tactile,
  title={Tactile-informed action primitives mitigate jamming in dense clutter},
  author={Brouwer, Dane and Citron, Joshua and Choi, Hojung and Lepert, Marion and Lin, Michael and Bohg, Jeannette and Cutkosky, Mark},
  booktitle={2024 IEEE International Conference on Robotics and Automation (ICRA)},
  pages={7991--7997},
  year={2024},
  organization={IEEE}
}

@inproceedings{zitkovich2023rt,
  title={Rt-2: Vision-language-action models transfer web knowledge to robotic control},
  author={Zitkovich, Brianna and Yu, Tianhe and Xu, Sichun and Xu, Peng and Xiao, Ted and Xia, Fei and Wu, Jialin and Wohlhart, Paul and Welker, Stefan and Wahid, Ayzaan and others},
  booktitle={Conference on Robot Learning},
  pages={2165--2183},
  year={2023},
  organization={PMLR}
}

@article{gansterer2014simulation,
  title={Simulation-based optimization methods for setting production planning parameters},
  author={Gansterer, Margaretha and Almeder, Christian and Hartl, Richard F},
  journal={International Journal of Production Economics},
  volume={151},
  pages={206--213},
  year={2014},
  publisher={Elsevier}
}

@INPROCEEDINGS{10161006,
  author={Saxena, Dhruv Mauria and Likhachev, Maxim},
  booktitle={2023 IEEE International Conference on Robotics and Automation (ICRA)}, 
  title={Planning for Complex Non-prehensile Manipulation Among Movable Objects by Interleaving Multi-Agent Pathfinding and Physics-Based Simulation}, 
  year={2023},
  volume={},
  number={},
  pages={8141-8147},
  doi={10.1109/ICRA48891.2023.10161006}}

@article{donald1993kinodynamic,
  title={Kinodynamic motion planning},
  author={Donald, Bruce and Xavier, Patrick and Canny, John and Reif, John},
  journal={Journal of the ACM (JACM)},
  volume={40},
  number={5},
  pages={1048--1066},
  year={1993},
  publisher={ACM New York, NY, USA}
}

@inproceedings{xiang2024grasping,
  title={Grasping trajectory optimization with point clouds},
  author={Xiang, Yu and Allu, Sai Haneesh and Peddi, Rohith and Summers, Tyler and Gogate, Vibhav},
  booktitle={2024 IEEE/RSJ International Conference on Intelligent Robots and Systems (IROS)},
  pages={9885--9892},
  year={2024},
  organization={IEEE}
}

@article{liu2022real,
  title={Real-time and efficient collision avoidance planning approach for safe human-robot interaction},
  author={Liu, Hongyan and Qu, Daokui and Xu, Fang and Du, Zhenjun and Jia, Kai and Song, Jilai and Liu, Mingmin},
  journal={Journal of Intelligent \& Robotic Systems},
  volume={105},
  number={4},
  pages={93},
  year={2022},
  publisher={Springer}
}

@article{braglia2023online,
  title={Online motion planning for safe human--robot cooperation using b-splines and hidden markov models},
  author={Braglia, Giovanni and Tagliavini, Matteo and Pini, Fabio and Biagiotti, Luigi},
  journal={Robotics},
  volume={12},
  number={4},
  pages={118},
  year={2023},
  publisher={MDPI}
}

@article{tang2024obstacle,
  title={Obstacle avoidance path planning of 6-DOF robotic arm based on improved A* algorithm and artificial potential field method},
  author={Tang, Xianxing and Zhou, Haibo and Xu, Tianying},
  journal={Robotica},
  volume={42},
  number={2},
  pages={457--481},
  year={2024},
  publisher={Cambridge University Press}
}

@article{odhner2014compliant,
  title={A compliant, underactuated hand for robust manipulation},
  author={Odhner, Lael U and Jentoft, Leif P and Claffee, Mark R and Corson, Nicholas and Tenzer, Yaroslav and Ma, Raymond R and Buehler, Martin and Kohout, Robert and Howe, Robert D and Dollar, Aaron M},
  journal={The International Journal of Robotics Research},
  volume={33},
  number={5},
  pages={736--752},
  year={2014},
  publisher={SAGE Publications Sage UK: London, England}
}

@article{zhou2024active,
  title={An active-passive compliance strategy for robotic plugging and unplugging of rocket electrical connectors},
  author={Zhou, Hao and Zhang, Xin and Liu, Jinguo and Ju, Zhaojie},
  journal={IEEE/ASME Transactions on Mechatronics},
  volume={30},
  number={2},
  pages={1014--1025},
  year={2024},
  publisher={IEEE}
}

@article{brooks2024video,
  title={Video generation models as world simulators},
  author={Brooks, Tim and Peebles, Bill and Holmes, Connor and DePue, Will and Guo, Yufei and Jing, Li and Schnurr, David and Taylor, Joe and Luhman, Troy and Luhman, Eric and others},
  journal={OpenAI Blog},
  volume={1},
  number={8},
  pages={1},
  year={2024}
}

@article{hang2021manipulation,
  title={Manipulation for self-identification, and self-identification for better manipulation},
  author={Hang, Kaiyu and Bircher, Walter G and Morgan, Andrew S and Dollar, Aaron M},
  journal={Science robotics},
  volume={6},
  number={54},
  pages={eabe1321},
  year={2021},
  publisher={American Association for the Advancement of Science}
}

@inproceedings{sundaresan2022diffcloud,
  title={Diffcloud: Real-to-sim from point clouds with differentiable simulation and rendering of deformable objects},
  author={Sundaresan, Priya and Antonova, Rika and Bohgl, Jeannette},
  booktitle={2022 IEEE/RSJ International Conference on Intelligent Robots and Systems (IROS)},
  pages={10828--10835},
  year={2022},
  organization={IEEE}
}

@inproceedings{ren2023kinodynamic,
  title={Kinodynamic rapidly-exploring random forest for rearrangement-based nonprehensile manipulation},
  author={Ren, Kejia and Chanrungmaneekul, Podshara and Kavraki, Lydia E and Hang, Kaiyu},
  booktitle={2023 IEEE International Conference on Robotics and Automation (ICRA)},
  pages={8127--8133},
  year={2023},
  organization={IEEE}
}

@article{dogar2012physics,
  title={Physics-based grasp planning through clutter},
  author={Dogar, M and Hsiao, Kaijen and Ciocarlie, Matei and Srinivasa, Siddhartha},
  journal={Robot.: Sci. Syst},
  pages={57--64},
  year={2012}
}

@inproceedings{yoshida2015simulation,
  title={Simulation-based optimal motion planning for deformable object},
  author={Yoshida, Eiichi and Ayusawa, Ko and Ramirez-Alpizar, Ixchel G and Harada, Kensuke and Duriez, Christian and Kheddar, Abderrahmane},
  booktitle={2015 IEEE international workshop on advanced robotics and its social impacts (ARSO)},
  pages={1--6},
  year={2015},
  organization={IEEE}
}

@book{pang2023planning,
  title={Planning, sensing, and control for contact-rich robotic manipulation with quasi-static contact models},
  author={Pang, Tao},
  year={2023},
  publisher={Massachusetts Institute of Technology}
}

@article{wang2024trafficability,
  title={Trafficability anticipation for quadruped robot in field operation},
  author={Wang, Chengjin and Zhang, Rui and Dong, Wenchao and Li, Tao and Jiang, Lei and Xu, Wei and Xu, Peng and Zhou, Yanmin and Zou, Meng},
  journal={Journal of Field Robotics},
  volume={41},
  number={4},
  pages={851--866},
  year={2024},
  publisher={Wiley Online Library}
}

@article{de2022insect,
  title={Insect-inspired AI for autonomous robots},
  author={de Croon, Guido CHE and Dupeyroux, JJG and Fuller, Sawyer B and Marshall, James AR},
  journal={Science robotics},
  volume={7},
  number={67},
  pages={eabl6334},
  year={2022},
  publisher={American Association for the Advancement of Science}
}

@article{yoshioka2024sensory,
  title={Sensory-motor circuit is a therapeutic target for dystonia musculorum mice, a model of hereditary sensory and autonomic neuropathy 6},
  author={Yoshioka, Nozomu and Kurose, Masayuki and Sano, Hiromi and Tran, Dang Minh and Chiken, Satomi and Tainaka, Kazuki and Yamamura, Kensuke and Kobayashi, Kenta and Nambu, Atsushi and Takebayashi, Hirohide},
  journal={Science advances},
  volume={10},
  number={30},
  pages={eadj9335},
  year={2024},
  publisher={American Association for the Advancement of Science}
}

@article{rao2024sensory,
  title={A sensory--motor theory of the neocortex},
  author={Rao, Rajesh PN},
  journal={Nature neuroscience},
  volume={27},
  number={7},
  pages={1221--1235},
  year={2024},
  publisher={Nature Publishing Group US New York}
}

@article{joshi2023dynamic,
  title={Dynamic synchronization between hippocampal representations and stepping},
  author={Joshi, Abhilasha and Denovellis, Eric L and Mankili, Abhijith and Meneksedag, Yagiz and Davidson, Thomas J and Gillespie, Anna K and Guidera, Jennifer A and Roumis, Demetris and Frank, Loren M},
  journal={Nature},
  volume={617},
  number={7959},
  pages={125--131},
  year={2023},
  publisher={Nature Publishing Group UK London}
}

@article{sintov2023simple,
  title={Simple kinesthetic haptics for object recognition},
  author={Sintov, Avishai and Meir, Inbar},
  journal={The International Journal of Robotics Research},
  volume={42},
  number={7},
  pages={537--561},
  year={2023},
  publisher={SAGE Publications Sage UK: London, England}
}

@article{zhou2023tacsuit,
  title={TacSuit: A wearable large-area, bioinspired multimodal tactile skin for collaborative robots},
  author={Zhou, Yanmin and Zhao, Jiangang and Lu, Ping and Wang, Zhipeng and He, Bin},
  journal={IEEE Transactions on Industrial Electronics},
  volume={71},
  number={2},
  pages={1708--1717},
  year={2023},
  publisher={IEEE}
}

@article{ding2014combined,
  title={Combined state and least squares parameter estimation algorithms for dynamic systems},
  author={Ding, Feng},
  journal={Applied Mathematical Modelling},
  volume={38},
  number={1},
  pages={403--412},
  year={2014},
  publisher={Elsevier}
}

@article{distefano2014comparison,
  title={A comparison of diagonal weighted least squares robust estimation techniques for ordinal data},
  author={DiStefano, Christine and Morgan, Grant B},
  journal={Structural Equation Modeling: a multidisciplinary journal},
  volume={21},
  number={3},
  pages={425--438},
  year={2014},
  publisher={Taylor \& Francis}
}

@article{zhong2022soft,
  title={Soft tracking using contacts for cluttered objects to perform blind object retrieval},
  author={Zhong, Sheng and Fazeli, Nima and Berenson, Dmitry},
  journal={IEEE Robotics and Automation Letters},
  volume={7},
  number={2},
  pages={3507--3514},
  year={2022},
  publisher={IEEE}
}

@article{schumacher2019introductory,
  title={An introductory review of active compliant control},
  author={Schumacher, Marie and Wojtusch, Janis and Beckerle, Philipp and Von Stryk, Oskar},
  journal={Robotics and Autonomous Systems},
  volume={119},
  pages={185--200},
  year={2019},
  publisher={Elsevier}
}

@article{skrinjar2018review,
  title={A review of continuous contact-force models in multibody dynamics},
  author={Skrinjar, Luka and Slavi{\v{c}}, Janko and Bolte{\v{z}}ar, Miha},
  journal={International Journal of Mechanical Sciences},
  volume={145},
  pages={171--187},
  year={2018},
  publisher={Elsevier}
}

@article{gilardi2002literature,
  title={Literature survey of contact dynamics modelling},
  author={Gilardi, Gianni and Sharf, Inna},
  journal={Mechanism and machine theory},
  volume={37},
  number={10},
  pages={1213--1239},
  year={2002},
  publisher={Elsevier}
}

@article{ding2023least,
  title={Least squares parameter estimation and multi-innovation least squares methods for linear fitting problems from noisy data},
  author={Ding, Feng},
  journal={Journal of Computational and Applied Mathematics},
  volume={426},
  pages={115107},
  year={2023},
  publisher={Elsevier}
}

@article{arslan2019sensor,
  title={Sensor-based reactive navigation in unknown convex sphere worlds},
  author={Arslan, Omur and Koditschek, Daniel E},
  journal={The International Journal of Robotics Research},
  volume={38},
  number={2-3},
  pages={196--223},
  year={2019},
  publisher={SAGE Publications Sage UK: London, England}
}

@inproceedings{tzes2022reactive,
  title={Reactive informative planning for mobile manipulation tasks under sensing and environmental uncertainty},
  author={Tzes, Mariliza and Vasilopoulos, Vasileios and Kantaros, Yiannis and Pappas, George J},
  booktitle={2022 International Conference on Robotics and Automation (ICRA)},
  pages={7320--7326},
  year={2022},
  organization={IEEE}
}

@misc{coumans2016pybullet,
  title={Pybullet, a python module for physics simulation for games, robotics and machine learning},
  author={Coumans, Erwin and Bai, Yunfei},
  year={2016}
}

@article{scordamaglia2025autonomous,
  title={Autonomous Tracked Vehicles Operating in Cluttered and Unknown Environments: A Networked Set-Theoretic Receding Horizon Control Strategy},
  author={Scordamaglia, Valerio and Ferraro, Alessia and Franz{\`e}, Giuseppe},
  journal={IEEE Transactions on Cybernetics},
  year={2025},
  publisher={IEEE}
}

\begin{IEEEbiography}[{\includegraphics[width=1in,height=1.25in,clip,keepaspectratio]{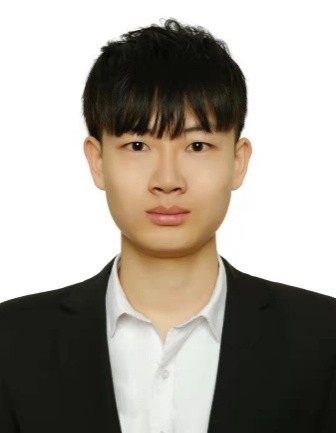}}]{Chengjin Wang (Student Member, IEEE)}  received the M.S. degree in the agricultural mechanization engineering from Jilin University, Changchun, China, in 2022. He is currently working toward a Ph.D. degree in intelligent science and technology at Tongji University, Shanghai, China.

His research interests lie at the intersection of robotics, computer graphics, and machine learning.
\end{IEEEbiography}

\begin{IEEEbiography}[{\includegraphics[width=1in,height=1.25in,clip,keepaspectratio]{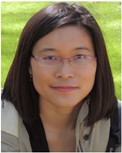}}]{Yanmin Zhou (Member, IEEE)} received the M.S. degree in control theory and control engineering from Tongji University, Shanghai, China, in 2011, and the Ph.D. degree in biomechanics from Cambridge University, Cambridge, U.K., in 2015.

She is currently a Vice Professor with the Department of Control Science and Engineering, College of Electronics and Information Engineering, Tongji University. Her current research interests include bionics, the design and control of intelligent robot.
\end{IEEEbiography}

\begin{IEEEbiography}[{\includegraphics[width=1in,height=1.25in,clip,keepaspectratio]{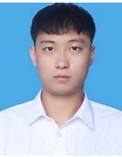}}]{Feng Luan} is currently working toward the Ph.D. degree in intelligent science and technology with the Shanghai Research Institute for Intelligent Autonomous Systems, Tongji University, Shanghai, China. 

His research interests include tactile sensors, 3-D reconstruction, and the design and control of intelligent robots.
\end{IEEEbiography}

\begin{IEEEbiography}[{\includegraphics[width=1in,height=1.25in,clip,keepaspectratio]{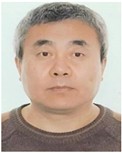}}]{Runjie Shen} received the B.S. degree from Xi'an Polytechnic University, Xi'an, China, in 1995, the M.S. degree from Xi'an Jiaotong University, Xi'an, in 1998, and the Ph.D. degree from Zhejiang University, Hangzhou, China, in 2004. 

From 2004 to 2006, he was a Post-Doctoral Researcher at Zhejiang University, where he became an Associate Researcher in 2006. Since 2011, he has been an Associate Researcher with Tongji University, Shanghai, China. His current research interests include sensing technologies and nondestructive.
\end{IEEEbiography}

\begin{IEEEbiography}[{\includegraphics[width=1in,height=1.25in,clip,keepaspectratio]{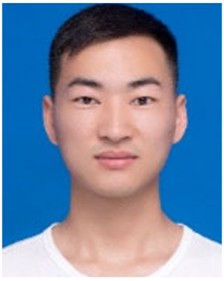}}]{Hongrui Sang}  received the B.E. degree in control technology and instruments from the University of Electronic Science and Technology of China (UESTC), Chengdu, China, in 2017, and the Ph.D. degree in control science and engineering from Tongji University, Shanghai, China, in 2024. 

He is currently a Lecturer with the School of Logistics Engineering, Shanghai Maritime University, Shanghai. His current research interests include reinforcement learning, visual navigation, and planning in unknown environment.
\end{IEEEbiography}

\begin{IEEEbiography}[{\includegraphics[width=1in,height=1.25in,clip,keepaspectratio]{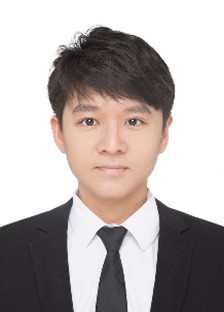}}]{Zheng Yan} separately received the B.E. degree the M.E. degree from Tianjin University (TJU), Tianjin, China, in 2019 and 2022. He is currently pursuing the Ph.D. degree in Intelligent Science and Technology with Shanghai Research Institute for Intelligent Autonomous Systems, Tongji University. 

His current research interests include human-robot interaction and tactile sensing based on electronic skin.
\end{IEEEbiography}

\begin{IEEEbiography}[{\includegraphics[width=1in,height=1.25in,clip,keepaspectratio]{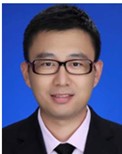}}]{Zhipeng Wang} received the M.S. degree in mechanical engineering from Zhejiang University, Hangzhou, China, in 2011, and the Ph.D. degree in mechanical engineering from Tongji University, Shanghai, China, in 2015.

Between 2015 and 2018, he held postdoc toral research appointments with the College of Mechanical Engineering, Tongji University. He is currently a Vice Professor with the Department of Control Science and Engineering, College of Electronics and Information Engineering, Tongji University. His current research interests include biped robot, mechatronics, and dynamics.
\end{IEEEbiography}

\begin{IEEEbiography}[{\includegraphics[width=1in,height=1.25in,clip,keepaspectratio]{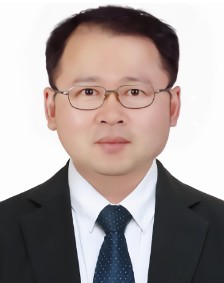}}]{Bin He (Senior Member, IEEE)} received the B.S. degree in engineering machinery from Jilin University, Changchun, China, in 1996, and the Ph.D. degree in mechanical and electronic control engineering from Zhejiang University, Hangzhou, China, in 2001. 

From 2001 to 2003, he was a Post-Doctoral Researcher with the State Key Laboratory of Fluid Power Transmission and Control. He is currently a Professor with the Department of Control Science and Engineering, College of Electronics and Information Engineering, Tongji University, Shanghai, China. His current research interests include intelligent robot control, biomimetic microrobots, and wireless networks.

\end{IEEEbiography}

\end{document}